\documentclass[11pt]{article}
\usepackage{coling2020}
\usepackage{times}
\usepackage{url}
\usepackage{latexsym}

\usepackage{times}
\usepackage{latexsym}
\usepackage{tabularx}
\usepackage{amssymb}
\usepackage{amsmath}
\usepackage{enumitem}
\usepackage{booktabs}
\usepackage{url}
\usepackage{color} 
\usepackage{verbatim}
\usepackage{calc}
\usepackage{mathtools}
\usepackage[draft]{todonotes}
\usepackage{soul}
\usepackage{tabularx}
\usepackage[utf8]{inputenc}
\usepackage{pbox}
\usepackage{color, colortbl}
\usepackage{wrapfig}

\title{Words are the Window to the Soul:\\ Language-based User Representations for Fake News Detection}

\author{Marco Del Tredici \\
  Amazon \\
  {\tt mttredic@amazon.com} \\\And
  Raquel Fern\'andez \\
  University of Amsterdam \\
  {\tt raquel.fernandez@uva.nl} \\}

\date{}

\begin{document}
\maketitle
\begin{abstract}
Cognitive and social traits of individuals are reflected in language use. 
Moreover, individuals who are prone to spread fake news online often share common traits. 
Building on these ideas, we introduce a model that 
creates representations of individuals on social media based only on the language they produce, and use them to detect fake news. 
We show that language-based user representations are beneficial for this task. 
We also present an extended analysis of the language of fake news spreaders, showing that its main features are mostly domain independent and consistent across two English datasets. 
Finally, we exploit the relation between language use and connections in the social graph to assess 
the presence of 
the Echo Chamber effect
 in our data. 

\end{abstract}

\section{Introduction}
\label{sec:introduction}

\blfootnote{
    %
    %
    %
    %
     \hspace{-0.65cm}  
     This work is licensed under a Creative Commons 
     Attribution 4.0 International Licence.
     Licence details:
     \url{http://creativecommons.org/licenses/by/4.0/}.
    %
    %
}

Fake news have become a problem of paramount relevance in our society, due to their large diffusion in public discourse, especially on social media, and their alarming effects on our lives \cite{lazer2018science}.
Several works show that fake news played a role in major events such as the US Presidential Elections \cite{allcott2017social}, stock market trends \cite{rapoza2017can}, and the Coronavirus disease outbreak  \cite{shimizu20202019}.
%
In NLP a considerable amount of work has been dedicated to fake news detection, i.e., the task of classifying a news as either real or fake -- see \newcite{zhou2018fake}, \newcite{kumar2018false} and \newcite{oshikawa2018survey} for overviews. 
While initial work  focused uniquely on the textual content of the news \cite{mihalcea2009lie}, subsequent research has considered also the social context in which news are consumed, characterizing, in particular, the
users who spread news in social media.
In line with the results reported in other classification tasks of user-generated texts \cite{del2019you,pan2019social}, several studies show that leveraging user representations, together with news' ones, leads to improvements in fake news detection.
In these studies, user representations are usually computed using informative but costly features, such as manually assigned credibility scores \cite{kirilin2018exploiting}. 
Other studies, which leverage largely available but scarcely informative features
(e.g., connections on social media), report less encouraging results \cite{zubiaga2016analysing}.

Our work also focuses on users. 
We build on psychological studies that show that some people are more prone than others to spread fake news, and that these people usually share a set of cognitive and social factors, such as personality traits, beliefs and ideology \cite{pennycook2015makes,pennycook2017falls}. 
Also, we rely on studies showing a relation between these 
factors and language use, both in Psychology and Linguistics  \cite{pennebaker2003psychological,de2012discourse} and
in NLP \cite{plank2015personality,preoctiuc2017beyond}.
We therefore propose to leverage user-generated language, an abundant resource in social media, to create user representations based solely on users' language production. We expect, in this way, to indirectly capture the factors characterizing people who spread fake news. 

We implement a model for fake news detection which jointly models news and user-generated texts. 
We use Convolutional Neural Networks (CNNs), which were shown to perform well on text classification tasks \cite{kalchbrenner2014convolutional} and are highly interpretable \cite{jacovi2018understanding}, i.e., they allow us to extract the informative linguistic features of the input texts.
We test our model on two public English datasets for fake news detection based on Twitter data, both including news and, for each news, the users who spread them on Twitter. 
We leverage two kinds of user-generated language, i.e., past tweets and self-description. In line with our expectations, model performance improves when language-based user representations are coupled with news representations, compared to when only the latter are used.
Moreover, the model achieves high results when leveraging user-generated texts only to perform the task.

We use the linguistic features returned by the model to 
analyze the language of fake news spreaders, showing that it has distinctive features related to both content, e.g., 
the large usage of words related to emotions and topics such as family and religion, 
and style, e.g., a peculiar usage of punctuation. 
Importantly, these features are largely independent from the domain of the dataset, and stable across datasets. 
Moreover, we find that the two kinds of user-generated language we consider provide partially overlapping information, but with some relevant differences.


  
Finally, we consider the relation between the language produced by the users and their connections in the social graph.
In particular, we investigate the Echo Chamber effect, i.e, the situation in which the ideas expressed by a user are reinforced by their connections  \cite{jamieson2008echo}. 
In previous NLP work, the effect has been studied
by observing whether users connected in the social graph post the same content, usually defined as a link to a web page from a manually compiled list 
\cite{garimella2018political,choi2020rumor}.
We propose to define the content produced by the users based on their linguistic production, 
and to compute the Echo Chamber effect as a function of the similarity between the content of connected users and their distance in the social graph.  
By applying our methodology, we show that the Echo Chamber effect is at play, to different extent, in both the datasets under scrutiny. 


Modelling user-generated data requires careful consideration of the possible ethical aspects related to the treatment of such data. We provide an ethics statement in Appendix \ref{appendix:a}
with details on how we have dealt with these aspects.

\section{Related Work}
\label{sec:related_work}

Several studies on fake news detection focus uniquely on the text of the news \cite{mihalcea2009lie,rashkin2017truth,perez2018automatic}.
Despite some positive results, this approach is inherently undermined by the fact that fake news are often written in such a way as to look like real news. 
More recently, researchers have explored the possibility to leverage social information together with the one derived from news texts.
Some works focus on the patterns of propagation of fake news in social networks.  
\newcite{vosoughi2018spread} show that, compared to real news, fake news have deeper propagation trees, spread faster and reach a wider audience, 
while 
\newcite{ma2017detect} show that fake news originate from users with a few followers, and are spread by influential people.
A parallel line of research considers the users who spread fake news.
Some works focus on the detection of non-human agents (\textit{bots}) involved in the spreading process \cite{bessi2016social},
while others model the characteristics of human spreaders, as we do in this paper. 
\newcite{gupta2013faking} and \newcite{zubiaga2016analysing} represent users with simple features such as longevity on Twitter and following/friends relations, and show that these features have limited predictive power.
\newcite{kirilin2018exploiting}, \newcite{long2017fake} and \newcite{reis2019supervised} use more informative features, such as users' political party affiliation, job and credibility scores.
While leading to improvements on the task, features of this kind are usually either hard to retrieve 
or have to be manually defined, 
which hinders the possibility to scale the methodology to large sets of unseen users. 
\newcite{guess2019less} and \newcite{shu2019beyond}
  rely on manually annotated lists of news providers,
   thus presenting a similar scalability problem.
Finally, \newcite{shu2019role} 
represent users by mixing different kinds of information, e.g., previous tweets, location, and profile image. 
This approach shares with ours the usage of the previous tweets of a user (as will be explained in Section~\ref{subsec:users}). 
However, to our knowledge, we are the first to create user representations based \textit{uniquely} on users' linguistic production. 


Previous NLP work showed the presence of the Echo Chamber effect (ECE) on social media, especially in relation to political discourse \cite{ul2019survey}.
The majority of the studies implement a similar approach, whereby the ECE is said to exist if users which are connected in the social graph post the same content.
Usually, the content considered in these studies is a link to a web page from an annotated list.
For example, \newcite{del2016spreading} investigate the relation between echo chambers and spread of conspiracy theories by observing users that share links to pages promoting this kind of theories. \newcite{choi2020rumor} apply a similar approach to the analysis of rumours spread, while other studies adopt it to investigate echo chambers in relation to political polarization, in which case, links are labelled with political affiliation \cite{colleoni2014echo,garimella2018political,gillani2018me}.
We adopt the same approach but, crucially, we define the shared content based on the linguistic production of the users.

\section{Data}
\label{sec:data}
 
\subsection{Datasets}
\label{subsec:datasets}


We use two datasets, PolitiFact and GossipCop, available in the data repository FakeNewsNet\footnote{\url{https://github.com/KaiDMML/FakeNewsNet}.} \cite{shu2018fakenewsnet}. 
While other datasets for fake news exist \cite{oshikawa2018survey}, those in FakeNewsNet provide the possibility to retrieve the previous linguistic production of the users, thus making them particularly suitable for our purposes. However, these datasets are not annotated with the features used by previous work to represent users (see Section \ref{sec:related_work}), and hence a direct comparison between the language-based user representations we propose and the ones obtained with existing methodologies is not possible.
Both PolitiFact and GossipCop consist of a set of news labelled as either fake or real. 
PolitiFact (PF) includes political news from the website \url{https://www.politifact.com/}, whose labels were assigned by domain experts. 
News in GossipCop (GC) are about entertainment, and are taken from different sources. The labels of these news were assigned by the creators of the data repository.
For each news in the datasets, its title and body are available,\footnote{The body of the news is not in the downloadable dataset files, but it can be obtained using the code provided by the authors.} together with the IDs of the tweets that shared the news on Twitter.
We tokenize titles and bodies, set a maximum length of 1k tokens for bodies and 30 tokens for titles, and define news as the concatenation of their title and body.
We remove words that occur less than 10 times in the dataset, and replace URLs and integers with placeholders.
We add the tag $<$CAP$>$ before any all-caps word in order to keep information about style, 
and then lowercase the text. 
Finally, we keep only news which are spread by at least one user on Twitter (more details in Section~\ref{subsec:users}). We randomly split each dataset in train/validation/test (80/10/10).
In Table \ref{tab:data_fake_news_dataset} we report the number of fake and real news per dataset after our preprocessing.

\begin{wraptable}{r}{5cm}
\centering \small
\begin{tabular}{lcccc}
\toprule
\bf  & \bf  fake & \bf real & \bf users & \bf  DE  \\ 
\midrule
\bf PF &  362 & 367 & 20.7k & 79\%  \\
\bf GC & 2.5k & 4.9k & 62.5k & 82\%  \\
\bottomrule
\end{tabular}
\caption{Statistics for each dataset after preprocessing: Number of \textbf{fake} and \textbf{real} news; number of \textbf{users}; percentage of users for which a self-description (\textbf{DE}) is available.}
\label{tab:data_fake_news_dataset}
\end{wraptable} 


\subsection{Users}
\label{subsec:users}

The only information about users that we leverage is the language they produce. We retrieve it as follows. 
First, for each news, we identify the users who posted the tweets spreading the news.\footnote{In order to identify users and retrieve their information, we query the Twitter API using the Python library \texttt{tweepy}.} For some news it is not possible to find any user, due to the fact that the tweets were cancelled or that the user is not on Twitter anymore. We remove these news from the datasets. 
Also, in both datasets there are some users who spread many news. One risk, in this case, is that the model may memorize these users, rather than focus on general linguistic features. For this reason we keep only unique users per news, i.e., users who spread only one news in the dataset.  
Finally, we randomly subsample a maximum of 50 users per news, in order to make the data computationally tractable. As a result, for each news we obtain a set including 1 to 50 users who retweet it (on average, 28 users per news for PF and 9 for GC).
For each of these users, we retrieve their \textit{timeline} (TL), i.e., the concatenation of their previous tweets, and their \textit{description} (DE), i.e., the short text where users describe themselves on their profile. 
We expect descriptions and timelines to provide different information, the former being a short text written to present oneself, while tweets are written to comment on events, express opinions, etc.  Note that the description is optional, and not all the users provide it.
We set a maximum length of 1k tokens for timelines and 50 tokens for descriptions, and we apply to both the same preprocessing steps detailed in Section \ref{subsec:datasets}. Additionally, we add the tag $<$EMOJI$>$ before each emoji.
In Table \ref{tab:data_fake_news_dataset} we report the number of users per dataset, and the percentage for which a description is available.

\section{Model}
\label{sec:model}

We implement a model which takes as input a news $n$ and the set  $U=\{u_1, u_2, ..., u_i\}$ of texts produced by the users that spread $n$, and classifies the news as either fake or real. 
The model consists of two modules, one for news and one for user-generated texts, both implemented using Convolutional Neural Networks (CNNs).
The two modules can be used in parallel or independently (see Section \ref{sec:experiments}).
%
The news module takes as input $n$ and computes a vector $\textbf{n} \in \mathbb{R}^{d}$, where $d$ is equal to the number of filters of the CNN (see below).
The users module takes as input $U$ and returns a vector $\textbf{u} \in \mathbb{R}^{d}$, which is the weighted sum of the representations computed for user-generated texts in $U$.\footnote{The fact that vectors \textbf{n} and \textbf{u} have equal dimensionality is not a constraint  of the model but a methodological choice.} 
Vectors $\textbf{n}$ and $\textbf{u}$ are weighted by a gating system which controls for their contribution to the prediction, and then concatenated. 
The resulting vector is fed into a one-layer linear classifier $\textbf{W}$ $\in \mathbb{R}^{d+d \times 2}$, where 2 is the number of output classes (real and fake), which returns the logits vector $\textbf{o} \in \mathbb{R}^{2}$, on which softmax is computed.\footnote{We report the details of the implementation in Appendix \ref{appendix:b}.}

\paragraph{Extracting Linguistic Features from CNNs}

Recently, model interpretability has gained much traction in NLP, and an increasing number of studies have focused on understanding the inner-workings and the representations created by neural models \cite{alishahi2019analyzing}. Inspired by this line of work, and, in particular, by the analysis of CNNs for text classification by \newcite{jacovi2018understanding}, we inspect our model in order to extract the linguistic features it leverages for the final prediction, which we use  for our analysis (see Section \ref{sec:linguistic_analysis}). 
We describe below how we extract the relevant linguistic features from the model.
 
A CNN consists of one or more convolutional layers, and  each layer includes a number of \textit{filters} (or kernels). 
Filters are small matrices of learnable parameters which \textit{activate} (i.e., return an activation value) on the n-grams in the input text which are relevant for the final prediction: The higher the activation value, the more important the n-gram is for the prediction.\footnote{The size of a filter corresponds to the length of the n-grams it activates on. Hence, a filter of size 2 activates on bi-grams.}
As a first step, we collect all the relevant n-grams returned by the filters in the model.
Then, we assess which n-grams are relevant for the fake class, and which for the real class.
We do this by considering the \textit{contribution} of each filter to the two target classes, which is defined by the parameters in $\textbf{W}$ $\in \mathbb{R}^{d+d \times 2}$ \cite{jacovi2018understanding}. The contribution of  filter $f$ to the real and fake classes is determined, respectively, by parameters $\textbf{W}_{f0}$ and $\textbf{W}_{f1}$: if the former is positive and the latter negative, we say that $f$ contributes positively to the real class, and, therefore, the n-grams detected by $f$ are relevant for that class.
Consequently, for n-gram $x$ returned by the filter $f$ with activation value $v$, we compute the importance of $x$ for the class real as $R_v = v \times \textbf{W}_{f0}$ and for the class fake as $F_v = v \times \textbf{W}_{f1}$.

\section{Experimental Setup}
\label{sec:experiments}
 

\paragraph{Setups and Baseline}
Our goal is to assess the contribution of language-based user representations to the task of fake news detection. 
Thus, for each dataset, we implement the following setups:

\begin{itemize}

\setlength\itemsep{0em}
\item[-] \textbf{News}: We assess model performance when only news information is available.

\item[-] \textbf{TL / DE / TL+DE}: We provide the model only with user information. User information can be either the timeline (TL), the description (DE) or their concatenation (TL+DE).

\item[-] \textbf{N+TL / N+DE / N+TL+DE}: The model is provided with combined information from both news (N) and user-generated texts, which can again be in the three variants defined above.
\end{itemize}
We implement a Support Vector Machine (SVM) \cite{cortes1995support}  as a baseline. SVMs have been shown to achieve results which are comparable to those by neural-based models on text classification tasks \cite{basile2018simply}, and we thus expect the model to be a strong baseline.


\paragraph{Hyperparameters}
For each setup we perform grid hyperparameter search on the validation set using early stopping with patience value 10.  
We experiment with values 10, 20 and 40 for the number of filters, and 0.0, 0.2, 0.4 and 0.6 for dropout. 
In all setups batch size is equal to 8, filters focus on uni-grams, bi-grams and tri-grams, and we use Adam optimizer \cite{kingma2015adam} with learning rate of 0.001, $\beta_1$ = 0.9 and $\beta_2$ = 0.999. 
All the CNN modules have depth 1, and are initialized with 200-d GloVe embeddings pretrained on Twitter \cite{pennington2014glove}. 

We train the SVM baseline on uni-grams, bi-grams and tri-grams.
When modelling user information, we concatenate the user-generated texts of the users spreading the target news.
We use the \texttt{rbf} kernel, and perform grid hyperparameter search on the validation set. We explore values 1, 2, 5, 10, 15 and 30 for the hyperparameter C, and $1^{e-05}$, $1^{e-04}$, $1^{e-03}$, $1^{e-02}$, 1.0 for $ \gamma$.

For both CNN and SVM models, we use binary F-score as optimization metric, and indicate the fake class as the target class.

\section{Results}
\label{sec:results}

\begin{table*}[t] 
\centering
\begin{tabular}{l|c|l|lll|lll}
\toprule
  \bf	Dataset		  &   \bf	Model	&   \multicolumn{1}{c|}{\bf News}	  &   		\multicolumn{3}{c|}{\bf User Information}              & \multicolumn{3}{c}{ \bf Combined Information}   	\\
\bf  & \bf  & \bf  & \bf TL  & \bf DE  & \bf TL+DE & \bf N+TL & \bf N+DE & \bf  N+TL+DE \\
 \hline
PolitiFact (PF) &   SVM & 0.839 & 0.654 & 0.714  & 0.673 & 0.654 & 0.686 & 0.682 \\
%
 & CNN  & 0.865$^{\ast}$ & 0.812 & 0.706  & 0.824 & 0.888$^{\ast\diamond}$ & 0.879$^{\ast}$ & 0.882$^{\ast\diamond}$\\
 \hline
GossipCop (GC) & SVM & 0.629 & 0.505 & 0.439  & 0.514 & 0.518 & 0.609 & 0.525 \\
%
 & CNN & 0.641$^{\ast}$ & 0.545 & 0.463 & 0.526 & 0.710$^{\ast\diamond}$  & 0.714$^{\ast\diamond}$ & 0.719$^{\ast\diamond}$ \\ 
   \bottomrule
\end{tabular}
\caption{Results on the test set (binary F-score), for all the setups in our experiment. 
Standard deviation is in range [0.01-0.02] for all CNN setups. 
We group setups in which only information from user-generated texts is used (\textbf{User Information}) and those in which news and user-generated texts are jointly modelled (\textbf{Combined Information}).
For CNN, we mark with $^{\ast}$ the results which significantly improve over setups in User Information, while $\diamond$ indicates a significant improvement over the \textbf{News} setup.}
\label{tab:results}
\end{table*}

We report the results of the fake news detection task in Table~\ref{tab:results}. 
The results of 
our CNN model are computed as the average of 5 runs with different random initialization of the best model on the validation set. 
For SVM, we report the single result obtained by the best model on the validation set.\footnote{While a direct comparison to previous studies using the same dataset is not possible due to the specific preprocessing we applied to the data (see Section~\ref{sec:data}), 
the reported results are in line with those in the literature -- see, e.g., \newcite{shu2019beyond}.}

CNN outperforms SVM in all the setups, except for one.\footnote{Both CNN and SVM outperform a random baseline which samples labels based on their frequency in the dataset, and which obtains an F-score of 0.33 in GC and 0.48 on PF.}  
The largest improvements are in the \textbf{TL} and \textbf{TL+DE} setups for PF and in all the Combined Information setups: Our intuition is that these improvements are due to the weighted sum of the user vectors and to the gating system of the CNN (see Section \ref{sec:model}), which allow the model to pick the relevant information when the set of user-generated texts is large and includes long texts,\footnote{Recall that, on average, there are 28 users per news in PF and 9 in GC (see Section \ref{subsec:users}).} and when news and user-generated texts are jointly modelled.

We then focus on the performance of the CNN in the different setups.
First, we observe that results in the \textbf{News} setup are significantly higher than those in the User Information setups.\footnote{We compute statistically significant differences between sets of results using the unpaired Welch’s $t$ test.} 
This was expected, as classifying a news based on its text is presumably easier than by using only information about users who spread it. 
Nevertheless, the results in the \textbf{TL} setup are surprisingly high,  especially in PF, which indicates that the language used in timelines is highly informative.
The results in the \textbf{DE} setup, both in PF and GC, are lower than those in \textbf{TL}. The two setups, however, cannot be directly compared, as descriptions are not available for all users (see Section \ref{subsec:users}). 
When we re-run the models in the User Information setups keeping only users with both timeline and description, we observe no statistically significant differences between the results in the \textbf{TL} and \textbf{DE} setups.
Lastly, we observe no significant improvement when we add descriptions to timelines (i.e., \textbf{TL+DE} and \textbf{N+TL+DE} do not improve over \textbf{TL} and \textbf{N+TL}, respectively). 
Finally, in all the Combined Information setups the performance of the model significantly improves compared to the \textbf{News} setup -- except for \textbf{N+DE} in PF, for which the improvement is not statistically significant.
When we substitute user vectors with random ones in the Combined Information setups, we observe no improvement over the \textbf{News} setup.

Overall the results confirm our initial hypothesis that leveraging user representations based only on the language produced by users is beneficial for the task of fake news detection. They also raise interesting questions related to what makes user-generated language informative, and which qualitative differences exist, if any, between timelines and descriptions. We  address these questions in the next section.

\section{Linguistic Analysis}
\label{sec:linguistic_analysis}

\begin{wrapfigure}{r}{7.5cm}\centering
\includegraphics[width=7.5cm]{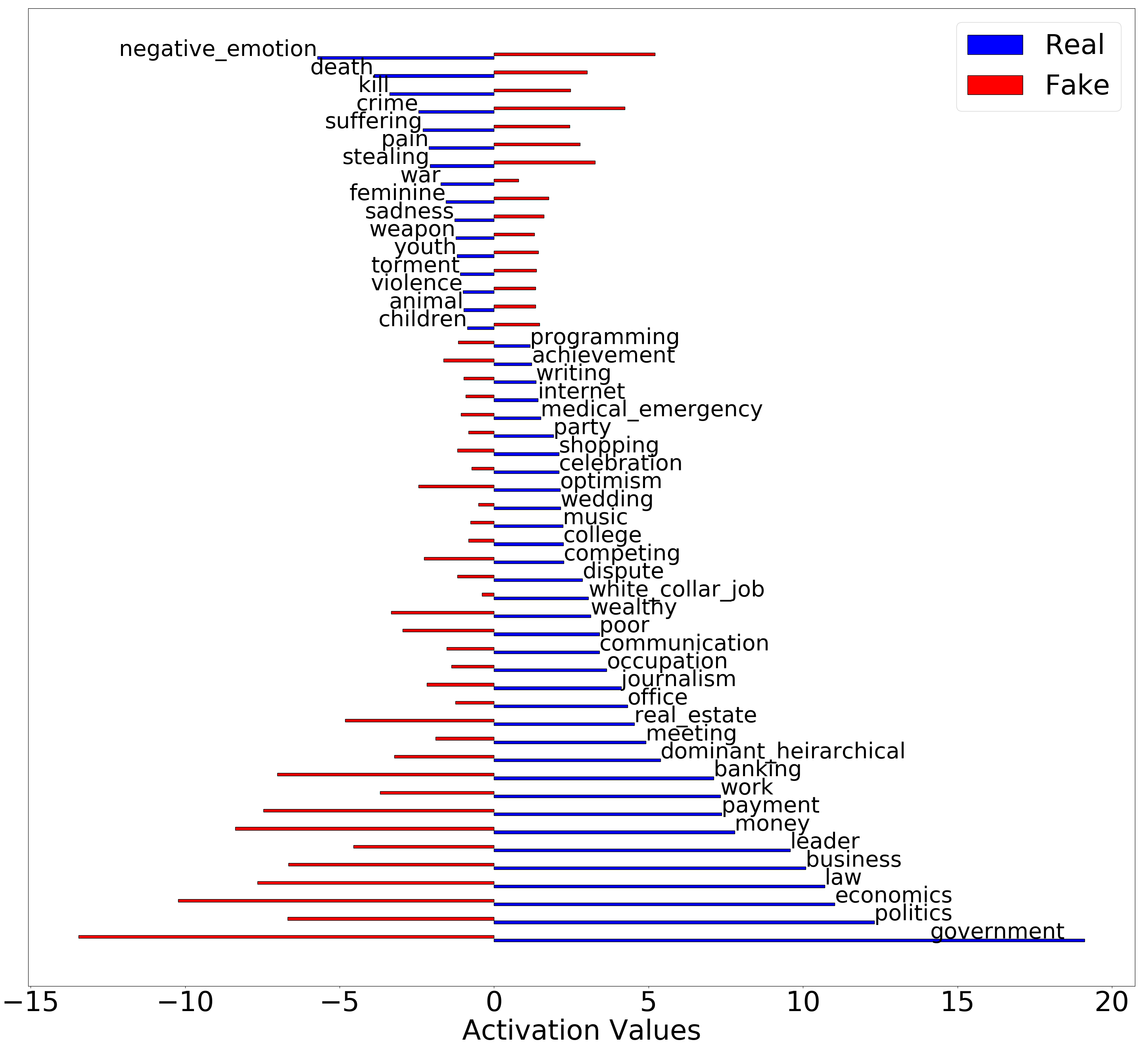}
\caption{Activation values of topics for the \textbf{News} setup in PF.
Best viewed in color.}
\label{fig:plot_ling_analysis}
\vspace*{-0.3cm}
\end{wrapfigure}

In this section, we analyse the language of news and of user-generated texts. We address two questions: 
(\textbf{Q1}) Which features of the language used by fake news spreaders are relevant for fake news detection, and how are they different from those of the language used by real news spreaders? 
(\textbf{Q2}) Which linguistic features do timelines and descriptions share, and which are different? Also, which features do these two kinds of user-generated texts share with the language of news?


To answer these questions, we need to analyse the language used in timelines, descriptions, and news independently.
We therefore consider, for both datasets, the models used at test time in \textbf{TL}, \textbf{DE} and \textbf{News}.
For each model, we extract the set of relevant n-grams, compute the $R_v$  and $F_v$ values for all of them, and sum the $R_v$  and $F_v$ of n-grams returned by more than one filter (see Section \ref{sec:model}). 
We use n-grams to analyse both style and content.
Regarding content, we analyse the \textbf{topic} of the n-grams, \textbf{proper names} and, for user-generated texts, \textbf{hashtags}. Regarding style, we consider  \textbf{punctuation marks}, \textbf{all-caps}, \textbf{function words} and, for user-generated texts, \textbf{emojis}.\footnote{We detect the topic using the Empath lexicon \cite{fast2016empath}, and use the LIWC lexicon \cite{pennebaker2001linguistic} to detect function words. We use the Python libraries \texttt{name-dataset} for proper names and  \texttt{emoji} for emojis.\label{fntools}} 
We check to which category, if any, each n-gram belongs to (e.g., \textit{trump} $\rightarrow$ \textbf{proper names} and \textit{\#usarmy} $\rightarrow$ \textbf{hashtags}).
The category \textbf{topic} includes a list of topics (e.g., Politics and War), and n-grams are assigned to these topics (e.g., \textit{missile}, \textit{army} $\rightarrow$ War). 
Similarly, \textbf{function words} includes several parts of speech (POS), hence, e.g., \textit{me}, \textit{you} $\rightarrow$ Pronouns. 
We define the importance of each topic and POS for the two target classes by summing the $R_v$ and $F_v$ values of the n-grams they include.
Finally, to consider only the n-grams which are relevant for one of the target classes, 
we compute the difference between $R_v$  and $F_v$ for each n-gram, compute the mean $\mu$ and standard deviation $\sigma$ of the differences, and keep only n-grams whose difference is larger than $\mu + \sigma$.

In Figure \ref{fig:plot_ling_analysis} we show the analysis of the topics for the \textbf{News} setup in PF. 
Red bars represent $F_v$ values, blue bars $R_v$ values:  
The higher the $R_v$ ($F_v$) value, the more the importance for the real (fake) class. For example, the topics Negative Emotions and Death are important for fake news; Government and Politics for real news.
Usually, to a large positive $F_v$ value  corresponds a large negative  $R_v$ value, and vice versa. 
We apply our methodology to address the questions introduced at the beginning of this section. 

\paragraph{Q1: The language of fake news spreaders}

\begin{wrapfigure}{l}{6.7cm}\centering
\includegraphics[width=6.7cm]{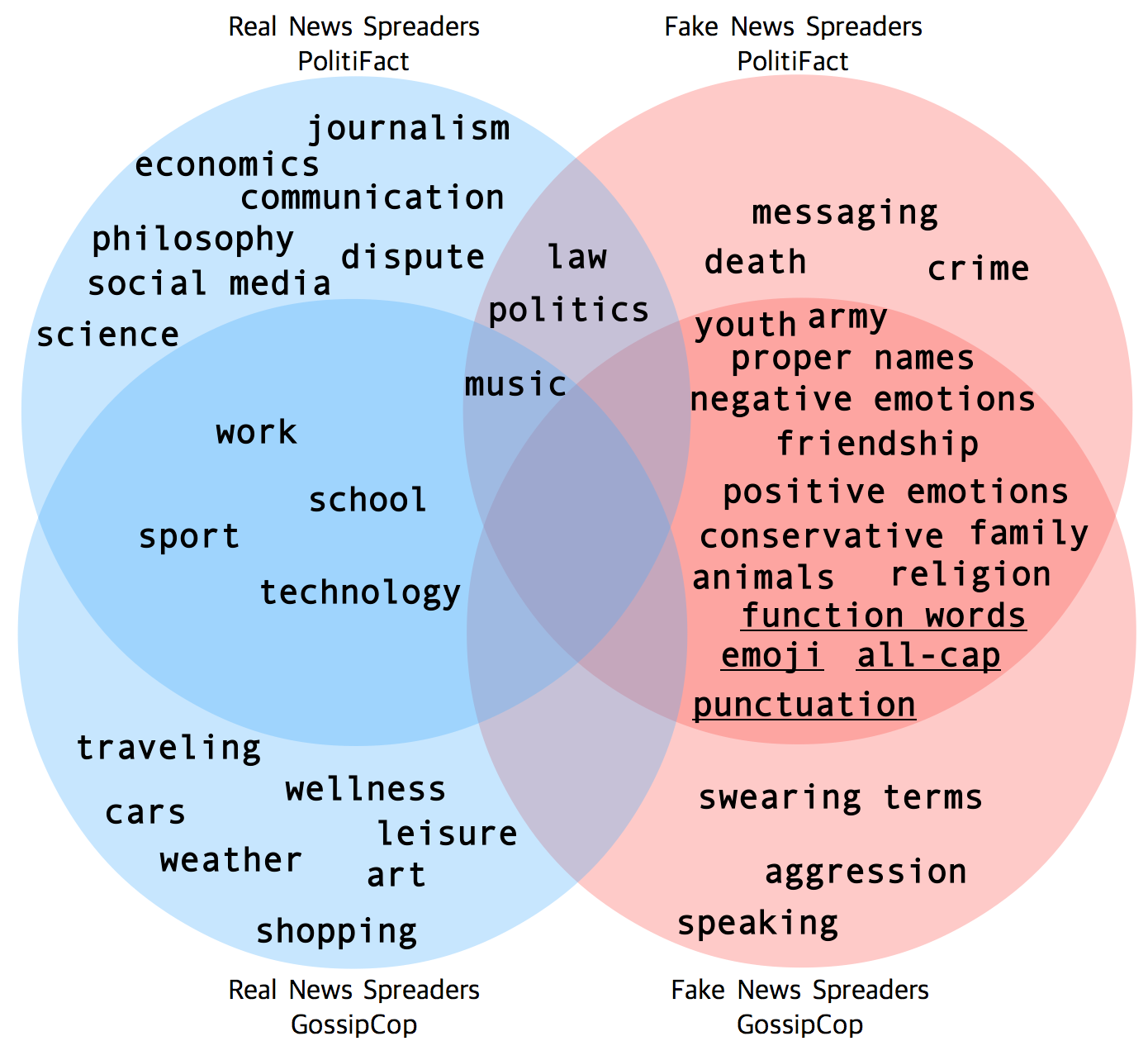}
\caption{The language of real news spreaders (blue circles) and fake news spreaders (red circles) in PF (top) and GC (bottom).}
\label{fig:q1}
\vspace*{0.1cm}
\includegraphics[width=6.7cm]{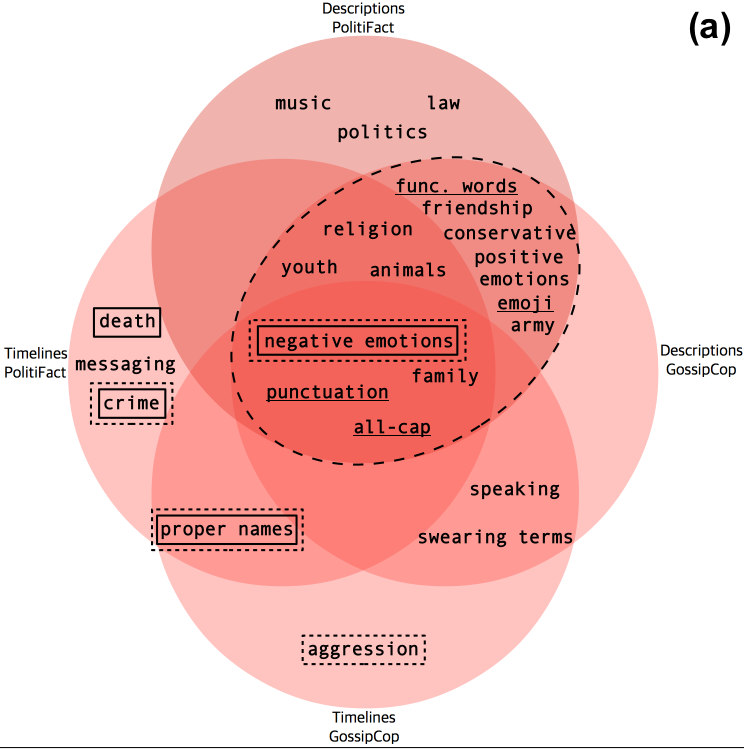}
\includegraphics[width=6.7cm]{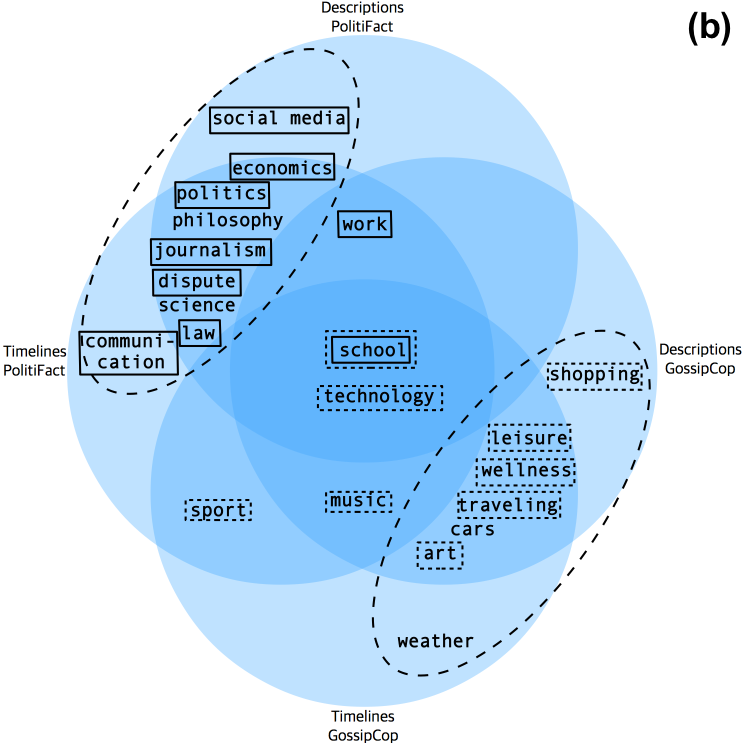}
\caption{
Relevant categories for TL and DE of fake news spreaders  (a) and real news spreaders (b). Solid-line boxes:  categories of fake (a) and real (b) news in PF. Dashed-line boxes:  categories of fake (a) and real (b) news in GC. }
\label{fig:q2}
\vspace*{-1cm}
\end{wrapfigure}

In Figure \ref{fig:q1} we show the main categories of the language of fake news spreaders (red circles) and real news spreaders (blue circles) in PF (top) and GC (bottom). Underlined categories refer to style, the others to content.\footnote{For simplicity, we aggregate similar topics, e.g., `positive emotions' includes topics such as Affection, Love and Optimism. 
}

A first observation is that very few categories are shared by the language of fake and real news spreaders (overlap between blue and red circles), and that those in common are mostly related to the domain of the dataset (e.g., law and politics in PF).
The language of fake news spreaders shows many common categories across datasets (overlap between red circles), mostly related to content. 
In particular, fake news spreaders of both datasets extensively talk about emotions and topics such as friendship, family, animals and religion. Interestingly, many of these topics are not directly related to the domain of either dataset.
The most important proper names (e.g., \textit{Jesus, Lord, Jehovah, Trump})  and hashtags (e.g., \textit{\#usarmy, \#trumptrain, \#god, \#prolife, \#buildthewall}) are again the same in the two datasets, and are highly related to the topics above. 
We observe some content-related categories which are not shared across datasets (non-overlapping areas in red circles), as they are related to the domain of the dataset (see \textbf{Q2}).
Cross-dataset consistency is even more evident for style: Fake news spreaders steadily use specific punctuation marks (quotes, hyphen, slash, question and exclamation mark), function words (first person pronouns and prepositions), emojis and words in all-caps.

The language of real news spreaders has different characteristics: many categories are dataset specific (non-overlapping areas in blue circles), while few of them are shared (overlap between blue circles). 
Also, dataset specific categories have higher activation values and are related to the domain of the dataset. Finally, no relevant style-related category is found for the language of real news spreaders.

Overall, the analysis shows that the language of fake news spreaders is clearly characterized by a set of linguistic features, related to both style and content.
Crucially, these features are largely domain-independent, and are consistently identified across datasets. This is in stark contrast with what is observed for the language of other users, which is more related to the domain of the dataset. 
These findings support the hypothesis that people who are more prone to spread fake news 
share a set of cognitive and sociological factors, which are mirrored in the features of the language they use.



\paragraph{Q2: The language of timelines, description and news}
We now analyse the relation between timelines, descriptions, and news. 
In Figure~\ref{fig:q2} we show the relevant categories of timelines and descriptions for fake (a) and real (b) news spreaders, in both datasets. 
The plots include the same information displayed in Figure~\ref{fig:q1}, but in greater detail.
In the plots, solid-line boxes indicate the relevant categories for the news shared by fake/real news spreaders in PF, dashed-line boxes the relevant categories for news shared by fake/real news spreaders in GC.

For fake news spreaders, we highlight the following findings.
First, the largest overlap (dotted ellipse) is observed between the descriptions \textit{across} the two datasets. 
Importantly, in this area we find the majority of categories which are not directly related to the domain of the datasets.
Second, in both datasets, timelines have some categories shared with descriptions (e.g., Negative Emotions and Punctuation), plus other categories related to the semantic field of violence (e.g., Crime and Aggression), together with Proper Names.
These timeline-specific categories are also the relevant ones for the fake news in PF (solid-line boxes) and in GC (dashed-line boxes). The relevance of similar categories across datasets is due to the fact that in both of them fake news are often built by mentioning a famous person (mainly Trump in PF, a celebrity in GC) in relation to some negative event -- a usual scheme in sensational news \cite{davis2003humans}.
In summary, all user-generated texts share some linguistic categories (central area of the plot), but it is in descriptions that we find the largest number of dataset-independent categories, related to both content and style, which characterize the language of fake news spreaders.
Conversely,  timelines share more categories with the news spread by the users.
These findings are in line with our expectations about the different nature of descriptions and timelines, as the former include more personal aspects of a user, while the latter are more related to the domain of the news they spread. 
Furthermore, the limited similarity between the language of fake news spreaders and 
of the news they spread provides further evidence to the hypothesis that the language of fake news spreaders is largely shaped by sociological and cognitive factors, and mostly independent from the domain.

For real news spreaders, 
there is a large overlap of content-related categories between timelines and descriptions \textit{within} a given dataset (dotted ellipses), while no style-related category is relevant for either kind of text. 
Differently from fake news spreaders, then, for real news spreaders descriptions and timelines do not present clear differences. 
Also, in both datasets, the relevant categories of real news 
strongly reflect the topics discussed in user-generated texts (see solid-line boxes for PF, and dashed-line boxes for GC).
We can thus conclude that a set of domain-related topics exists in each dataset, and that these topics are the relevant linguistic categories in timelines, description, and in news. In contrast, these texts do not share any characteristic related to style.  
\section{Echo Chamber Effect}
\label{sec:social_analysis}

\begin{wrapfigure}{R}{7.5cm}\centering
\includegraphics[width=7.5cm]{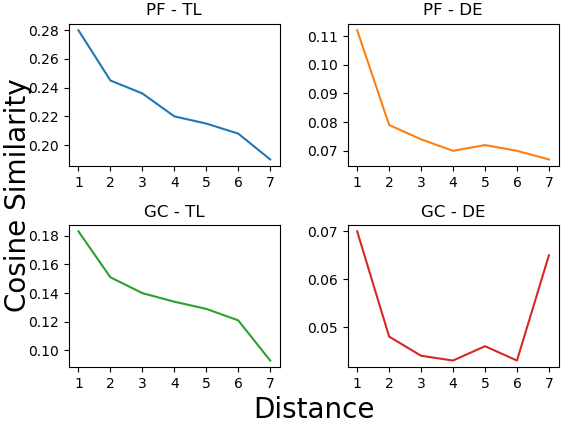}
\caption{The similarity values obtained for the two datasets with the \textit{TL-topic} vectors (TL) and with the \textit{DE-topic} vectors (DE).}
\label{fig:plots_ec}
\end{wrapfigure}
After showing the informativeness of language-based user representations, we now use them together with the information from the social graph to investigate the Echo Chamber effect (ECE).
We adopt the operational definition by  \newcite{garimella2018political}, and say that the ECE
exists when users in a social graph mostly receive content from their  connections which is similar to the content they produce.
We introduce a methodology to define the content produced by a user based on their language use, and to compute the ECE as a function of the content similarity of connected users and their distance in the social graph.

\paragraph{Social Graph}
\label{sec:social_graph}

To define the social graph we follow a common approach in the literature \cite{yang2017overcoming,del2019you} and create, for each dataset, a graph $G = (V,E)$ in which $V$ is the set of users in the dataset, and $E$ is the set of edges between them. An unweighted and undirected edge is instantiated between two users if one retweets the other. We retrieve information about retweets in users' timeline (see Section \ref{subsec:users}). 
In order to make the social graph more connected, we also add as nodes users who are not in the dataset, but have been retweeted at least 20 times by users in the dataset. The resulting graph for PF includes 32K nodes and 1.6M edges (density= 0.0031), the one for GC includes 109K nodes and 4.9M edges (density=0.0008).

\paragraph{User Representations}
\label{sec:user_representations}

To represent users based on their linguistic production, we adopt an approach similar to the one of Section~\ref{sec:linguistic_analysis}, and we first retrieve, for each user, the set of relevant n-grams and  their activation values.\footnote{In this case we ignore the class the n-gram is relevant for (i.e., the $R_v$ and $F_v$ values), and only consider value $v$ (see Section \ref{sec:model}).}
Since the ECE is related to the content posted by users, we consider only the topic of the n-grams, and ignore their style.\footnote{We do not consider proper names and hashtags because  the dimensionality of the resulting user vectors would be intractable.} 
Thus, for each user, we analyse the topic in their set of n-grams using again the Empath lexicon (see footnote~\ref{fntools}), 
 and we define a topic vector $t \in \mathbb{R}^{d}$, where $d$ is the number of topics in the lexicon,
  and $t_i$ is the activation value of the $i$-th topic.
We create two topic vectors per user, one based on the timeline (\textit{TL-topic}) and one on the description (\textit{DE-topic}), using the best models at test time in the \textbf{TL} and \textbf{DE} setups (see Section \ref{sec:experiments}).

\paragraph{Computing the Echo Chamber Effect}
\label{sec:compute_ec}

%
%
%

We conjecture that the ECE  exists for a user if the cosine similarity between their topic vector and the one of their connections \textit{decreases} as the distance (i.e, the number of hops away in the graph) \textit{increases}. 
To check the effect for all users in the graph, for each distance value, 
we compute the average cosine similarity of the users at that distance.\footnote{We consider distance values for which there are at least 100 connections. This results in a maximum distance of 7 for all social graphs.}

As shown in Figure \ref{fig:plots_ec}, we observe a monotonic decrease in similarity (Spearman $\rho$ $\leq$ $-0.9$, $p\!<\!0.005$) in all setups, except for GC-DE, where the decrease in similarity is much less pronounced and, consequently, the descending curve is more subject to fluctuations -- see the increase after distance 6.
However, for all setups there is significative negative difference between sets of values at consecutive distances (i.e., 1 and 2, 2 and 3, and so on) up to distance 4 (Welch’s $t$ test $p\!<\!0.005$). 
We believe that, overall, these results indicate that the ECE is present in our data, with different strength depending on the setup.
We also make the following observations.

First, we observe no  difference, in terms of ECE, between fake news and real news spreaders. This indicates that the effect is common to all users in the datasets, and not related to the cognitive and social traits which influence the language production of fake news spreaders (see Section~\ref{sec:linguistic_analysis}).



Second, in all the setups, the largest drop in similarity is observed between values at distances 1 and 2 or 2 and 3. 
We interpret this fact as an indication that the ECE is mostly at play, in our data, at close proximity. 
This result is in line with previous findings in Sociolinguistics which show that, in social networks, there are cliques of users linked by first or second order connections who mutually reinforce their ideas and practices \cite{labov1972,milroy1987language}.

As we inspect the results for timelines and descriptions, we observe that the former show higher similarity values on average, while the drop in similarity at distance 2/3 is more evident for descriptions. These findings are related to what observed in Section \ref{sec:linguistic_analysis}, as timelines share more domain-related topics, which causes them to be more similar to each other, while descriptions include more personal aspects, presumably shared with close connections. 


Finally,
the similarity values for both \textbf{TL} and \textbf{DE} are higher in PF than in GC. 
We believe this is due to the polarization of political groups in social networks, 
 whereby users belonging to the same political party tend to group in segregated clusters, with few external connections \cite{conover2011political}. 

\section{Conclusion}
\label{sec:conc}

In this work we 
addressed the task of fake news detection, and showed that results can be improved by leveraging user representations based \textit{uniquely} on the language the users produce. 
This improvement is due to the fact that the language used by fake news spreaders has specific and consistent features, which are captured by our model, and which we highlight in our analysis. 
Language-based user representations also allowed us to show the presence of the Echo Chamber effect in both our datasets.

Our results offer empirical confirmation of previous findings regarding the relation between language use and cognitive and social  factors, and they could foster further theoretical and computational work in this line of research. 
In particular, future computational work might address some of the limitations of the current study: 
For example, while we focus only on users spreading a single news, it would be interesting to model also users who spread multiple news, which are possibly both real and fake. 
Similarly, it would be relevant to investigate the \textit{cold start} problem, that is, the number of posts needed to create a reliable representation of the user, which is particularly important for newly registered users and for those who are not highly active.
Also, since the relation between language use and cognitive and social factors holds in every sphere of linguistic production, a natural extension of this work would be to apply the same methodology to other tasks involving user-generated language, such as, for example, suicidal prevention and mental disorders detection. 
Finally,
we hope the tools and insights provided in this study might be used 
to fight the diffusion of fake news, for example, by identifying and warning users who are vulnerable to them.

%

\section*{Acknowledgements}

This research has received funding from the Netherlands Organisation for Scientific Research (NWO) under VIDI grant nr. 276-89-008, \textit{Asymmetry in Conversation}. We thank the anonymous reviewers for their comments as well as the area chairs and PC chairs of COLING 2020.
The work presented in this paper was entirely conducted when the first author was affiliated with the University of Amsterdam, prior to working at Amazon.



\bibliography{coling2020}
\bibliographystyle{acl}

\appendix
\section{Appendix: Ethics Statement}
\label{appendix:a}

We begin by clarifying our motivation for this work. 
We build on studies showing that,
while there are malicious users who consciously spread fake news for different (usually unethical) reasons, others do so simply because they are not able to distinguish them from real news~\cite{pennycook2017falls,kumar2018false}. 
Our goal is to implement a system which helps to automatically identify these vulnerable users, not to hold them up to public disdain but, rather, to warn them of the risk to be involuntarily involved in a harmful process.

Nowadays, many studies in NLP focus on tasks related to concrete societal issues, for example, hate or abusive speech detection, suicidal prevention, or 
fake news detection as we do here. 
This line of research leverages user-generated data extracted from online social media.
This raises the question of how these sensitive data should be managed.
Several studies have been concerned with the ethical treatment of user-generated data, both in NLP and related fields \cite{vitak2016beyond,leidner2017ethical,schmaltz2018utility,olteanu2019social}, focusing on different aspects and proposing good practices.
We did our best to follow such practices. Concretely:

\begin{itemize}[leftmargin=13pt]
\item We collected and used only data made publicly available by the users, that we obtained using the Twitter API. In this case, then, no approval and informed consent from the users were  needed.
\item The data only include users and tweet IDs: In no case did we try to trace these IDs back to the real identity of the users.
\item We controlled for possible biases in our data processing. For example, we randomly sub-sampled users and we applied the same pre-processing to all user-generated content, as described in Section~\ref{sec:data}.
\item We do not derive any conclusion about specific users or groups of users. Rather, we focus our attention on language use, and its connections to psychological and societal factors.
\end{itemize}

\section{Appendix: Model}
\label{appendix:b}

We provide here a more detailed description of the model used in our experiments.
The input of the  model are the news $n$ and the set  $U=\{u_1, u_2, ..., u_i\}$ of texts produced by the users that spread $n$.
The model has two modules, one for news and one for user-generated texts, which can be used in parallel or independently.
The news module takes as input $n$ and computes vector $\textbf{n} \in \mathbb{R}^{d}$, where $d$ is equal to the number of filters of the CNN.
The users module takes as input $U$ and initially computes the matrix $\textbf{U} \in \mathbb{R}^{m,d}$, where $m$ is the number of users in $U$, and vector $\textbf{u}_i \in \mathbb{R}^{d}$ represents user $u_i$ in set $U$.
We assume not all the users to be equally relevant for the final prediction, and we therefore implement a gating system as linear layer $\textbf{W}_g$ $\in \mathbb{R}^{d \times 1}$, which takes as input $\textbf{U}$ and returns the vector $s \in \mathbb{R}^{m} $. A sigmoid function is applied to $s$, squeezing the values in it in range [0-1], where 0 means that the information from a user-generated text is not relevant, and 1 that it is maximally relevant.  The matrix of the weighted representations of the users is thus obtained as $\textbf{U'} = \textbf{U} \times s$. We finally compress user information in a single vector $\textbf{u} \in \mathbb{R}^{d}$ computed as $\textbf{u} = \sum\limits_{i=1}^m \textbf{u}_i \in \textbf{U'}$.
Vectors $\textbf{n}$ and $\textbf{u}$ are weighted by a gating system which controls for their contribution to the prediction, concatenated, and fed into a one-layer linear classifier $\textbf{W}$ $\in \mathbb{R}^{d+d \times 2}$, where 2 is the number of output classes (real and fake), which returns the logits vector $\textbf{o} \in \mathbb{R}^{2}$, on which softmax is computed.

\end{document}